\documentclass[runningheads]{llncs}
\usepackage{graphicx}
\usepackage{amsmath,amssymb} % define this before the line numbering.
\usepackage{color}
\usepackage{cite}
\usepackage{times}
\usepackage{soul}
\usepackage{multirow}
\usepackage{booktabs}
\usepackage{subcaption}
\usepackage[utf8]{inputenc}
\usepackage[width=122mm,left=12mm,paperwidth=146mm,height=193mm,top=12mm,paperheight=217mm]{geometry}
\usepackage{array}
\newcolumntype{L}[1]{>{\raggedright\let\newline\\\arraybackslash\hspace{0pt}}m{#1}}
\newcolumntype{C}[1]{>{\centering\let\newline\\\arraybackslash\hspace{0pt}}m{#1}}
\newcolumntype{R}[1]{>{\raggedleft\let\newline\\\arraybackslash\hspace{0pt}}m{#1}}

\begin{document}
\pagestyle{headings}
\mainmatter
\def\ECCV18SubNumber{21}  % Insert your submission number here

\title{Learning a Text-Video Embedding from \\ Incomplete and Heterogeneous Data} % Replae

%\titlerunning{ECCV-18 submission ID \ECCV18SubNumber}

\authorrunning{Antoine Miech, Ivan Laptev and Josef Sivic}

\author{Antoine Miech$^{1,2}$ \qquad Ivan Laptev$^{1,2}$ \qquad Josef Sivic$^{1,2,3}$}
\institute{$^{1}$École Normale Supérieure\thanks{$^1$WILLOW project, Département d’informatique, École Normale Supérieure, CNRS, PSL Research University, 75005 Paris, France.} \qquad $^{2}$Inria \qquad $^{3}$CIIRC\thanks{$^3$Czech Institute of Informatics, Robotics and Cybernetics at the Czech Technical University in Prague.} \\ \url{http://www.di.ens.fr/willow/research/mee/}}

\maketitle

\begin{abstract}

Joint understanding of video and language is an active research area with many applications.
Prior work in this domain typically relies on learning text-video embeddings.
One difficulty with this approach, however, is the lack of large-scale annotated video-caption datasets for training.
To address this issue, we aim at learning text-video embeddings from heterogeneous data sources.
To this end, we propose a Mixture-of-Embedding-Experts (MEE) model with ability to handle missing input modalities during training.
As a result, our framework can learn improved text-video embeddings simultaneously from image and video datasets.
We also show the generalization of MEE to other input modalities such as face descriptors.
We evaluate our method on the task of video retrieval and report results for the MPII Movie Description and MSR-VTT datasets.
The proposed MEE model demonstrates significant improvements and outperforms previously reported methods on both text-to-video and video-to-text retrieval tasks. Code: \ \url{https://github.com/antoine77340/Mixture-of-Embedding-Experts}

%\keywords{Video, Text, Multi-Modality, Retrieval, Joint-embedding, Missing data}
\end{abstract}

\section{Introduction}

Automatic video understanding is an active research topic with a wide range of applications including activity capture and recognition, video search, editing and description, video summarization and surveillance.
%Video understanding has recently gained significant attention in the research community. Being able to automatically understand videos leads to a vast amount of useful and practical applications such as: video filtering, organizing and classifying video databases, performing video summarization, video retrieval or video surveillance.
In particular, the joint understanding of video and natural language holds a promise to provide a convenient interface and to facilitate access to large amounts of video data.
Towards this goal recent works study representations of vision and language addressing tasks such as visual question answering~\cite{tapaswi16movieqa,yu17endtoend}, action learning and discovery~\cite{miech17learning,bojanowski15weakly,alayrac16unsupervised}, text-based event localization~\cite{hendricks17localizing} as well as video captioning, retrieval and summarization~\cite{pan16jointly,plummer2017enhancing,xu2015jointly,yu2016video}.
%One way to facilitate the communication between automatic machine video understanding and humans is to bridge the gap between natural language and videos. A significant amount of prior works addressed the problem of learning jointly from videos and natural language~\cite{klein15associating,miech17learning,pan16jointly,tapaswi16movieqa,yu17endtoend,bojanowski15weakly,hendricks17localizing,yu2016video,xu2015jointly}.
Notably, many of these works adopt and learn joint text-video representations where semantically similar video and text samples are mapped to close points in the {\em joint embedding space}.
Such representations have been proven efficient for joint text-video modeling e.g., in~\cite{pan16jointly,plummer2017enhancing,xu2015jointly,bojanowski15weakly,hendricks17localizing}.
%where video and text should be close in that space if and only if they are semantically similar.
%Indeed, learning a joint embedding space for video and text allows several practical applications such as:  Text-to-Video retrieval, Video-to-Text retrieval or video caption generation. 
%In this work, we also address the same problem of learning a joint Text-Video embedding space.

Learning video representations is known to require large amounts of training data \cite{tran15c3d,carreira2017quovadis}.
While video data with label annotations is already scarce, obtaining a large number of videos with text descriptions is even more difficult.
%When tackling this task, one shortcoming is the scarcity of large-scale annotated video-caption datasets.
Currently available video datasets with ground truth captions include DiDeMo~\cite{hendricks17localizing} (27K unique videos), MSR-VTT\cite{xu16msrvtt} (10K unique videos) and the MPII Movie Description dataset~\cite{rohrbach15dataset} (120K unique videos).
To compensate for the lack of video data, one possibility would be to pre-train visual representations on still image datasets~\cite{carreira2017quovadis} with object labels or image captions such as ImageNet~\cite{deng2009imagenet}, COCO~\cite{lin14coco}, Visual Genome~\cite{visualgenome} and Flickr30k~\cite{plummer2015flickr30k}. Pre-training, however, does not provide a principled way of learning from different data sources and suffers from the ``forgetting effect'' where the knowledge acquired from still images is removed during fine-tuning on video tasks. More generally, it would be beneficial to have methods that can learn embeddings simultaneously from heterogeneous and partially-available data sources such as appearance, motion and sound but also from other modalities such as facial expressions or human poses.

%Indeed, collecting and annotating large-scale video datasets has always been a cumbersome task in research. 
%Video-Caption datasets are no exception to this rule. On the other hand, it is less burdensome to collect and annotate large-scale image-caption datasets. Several large-scale image-caption datasets such as: COCO~\cite{lin14coco}, Visual Genome~\cite{visualgenome} and Flickr30k~\cite{plummer2015flickr30k} are available. It is natural to wonder if we can leverage an image-caption dataset to actually improve the learning of a Text-Video joint embedding. To the best of our knowledge, transfer learning via pre-training on images and fine-tuning on videos~\cite{carreira2017quovadis} is the main approach used to leverage image datasets to improve video models. The shortcomings with this approach are two-fold. First it does not leverage specifically image-caption dataset. Moreover the second stage of separately fine-tuning on videos has the effect of \textit{erasing} the knowledge previously acquired on images. In this work, we aim at going beyond these shortcomings by jointly learning on image-caption and video-caption datasets.

\begin{figure}[t]
  \begin{center}
    \mbox{}\vspace{.2cm}\\
     \includegraphics[width=\textwidth]{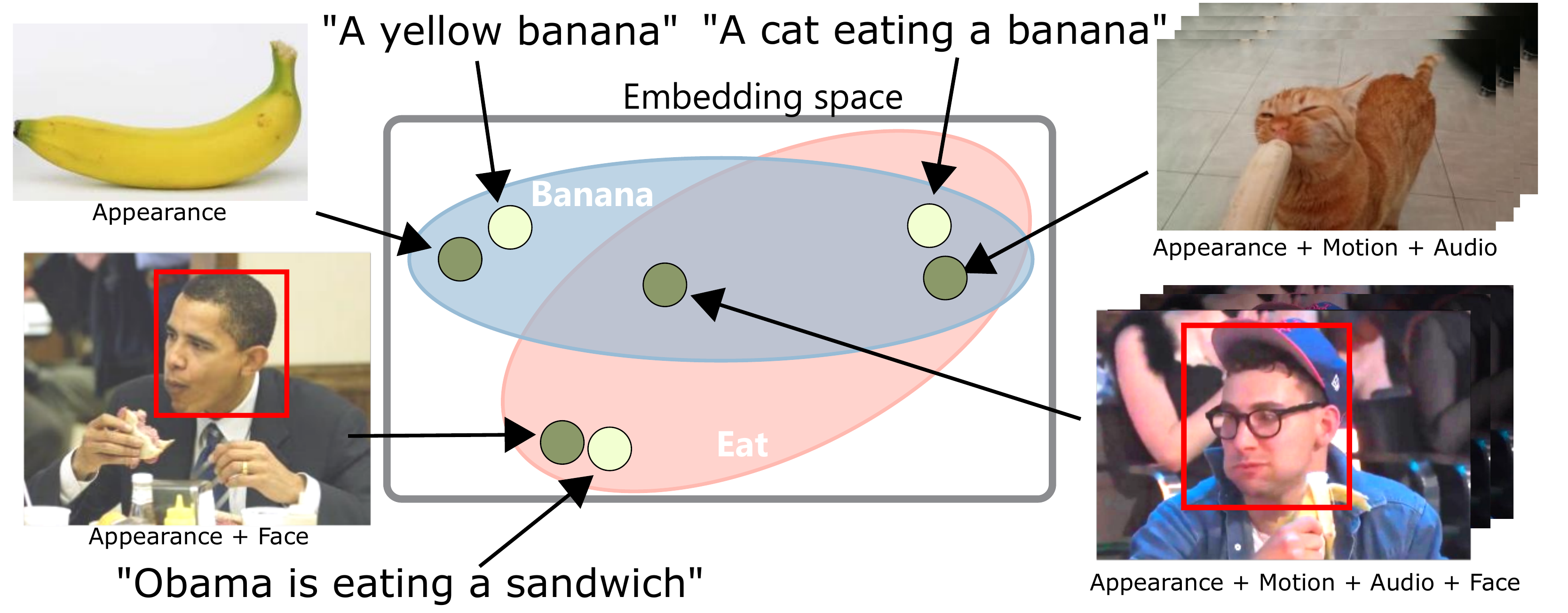}
\end{center}
\caption{\small We learn a text-video embedding from diverse and partially available data sources. This example illustrates a joint text-video embedding trained from videos and images while combining descriptors for global appearance, motion, audio and faces. The key advantage of our method is the ability to combine samples with different subsets of modalities, e.g., images with no motion and videos with no faces or sound.}
%\mbox{}\vspace{-1.0cm}\\
\label{fig:teaser}
\end{figure}

In this work we address the challenge of learning from heterogeneous data sources.
Our method is designed to learn a joint text-video embedding and is able to handle missing video modalities during training.
To enable this property, we propose a Mixture-of-Embedding-Experts (MEE) model that computes similarities between text and a varying number of video modalities.
The model is learned end-to-end and generates expert weights determining individual contributions of each modality.
During training we combine image-caption and video-caption datasets and treat images as a special case of videos without motion and sound.
For example, our method can learn an embedding for ``Eating banana'' even if ``banana'' only appears in training images but never in training videos (see Fig.~\ref{fig:teaser}).
We evaluate our method on the task of video retrieval and report results for the MPII Movie Description and MSR-VTT datasets.
The proposed MEE model demonstrates significant improvements and outperforms all previously reported methods on both text-to-video and video-to-text retrieval tasks.

Our MEE model can be easily extended to other data sources beyond global appearance, motion and sound.
In particular, faces in video contain valuable information including emotions, gender, age and identities of people.
As not all videos contain people, faces constitute a typical case of a potentially missing data source for our model.
To demonstrate the generalization of our model and to show the importance of faces for video retrieval, we compute facial descriptors for images and videos with faces.
We then treat faces as an additional data source in the MEE model and aggregate facial descriptors within a video (see Fig.~\ref{fig:separate_embd}).
The resulting MEE combining faces with appearance, motion and sound produces consistent improvements in our experiments.

\subsection{Contributions}
This paper provides the following contributions: \textit{(i)} First, we propose a new model for learning a joint text-video embedding called Mixture-of-Embedding-Experts (MEE).
The model is designed to handle missing video modalities during training and enables simultaneous learning from heterogeneous data sources.
%The main technical novelty of our approach makes it possible to leverage heterogeneous sources of data.
%To do so, our model is designed to properly handle missing video modalities during training.
\textit{(ii)} We showcase two applications of our framework. First, we can data augment video-caption datasets with image-caption datasets during training. We can also leverage face descriptors in videos to improve the joint text-video embedding. In both cases, we show improvements in several video retrieval benchmarks.
\textit{(iii)} By using MEE and leveraging multiple sources of training data we outperform state-of-the-art on the standard text-to-video and video-to-text retrieval benchmarks defined by the LSMDC~\cite{rohrbach15dataset} challenge.

\section{Related work}
In this section we review prior work related to vision and language, video representations and learning from sources with missing data.

\subsection{Vision and Language}
There is a large amount of work leveraging language in computer vision. Language is often used as a more powerful and subtle source of supervision than predefined classes. One way to leverage language in vision is to find a joint embedding space for both visual and textual modalities~\cite{gong14multi,klein15associating,pan16jointly,plummer2017enhancing,xu2015jointly,wang2016learning,wang2018learning,wu2017sampling}. In this common embedding space, visual and textual samples are close if and only if they are semantically similar. This common embedding space enables multiple applications such as text-to-image/video retrieval and image/video-to-text retrieval. The work of Aytar \textit{et al.}~\cite{aytar17see} is going further by learning a cross-modal embedding space for visual, textual and aural samples. In vision, language is also used in captioning where the task is to generate a descriptive caption of an image or a video\cite{johnson16densecap,pan16hierarchical,you16image,yu2016video}. Another related application is visual question answering~\cite{fukui16multimodal,malinowski15ask,tapaswi16movieqa,yu17endtoend}. A useful application of learning jointly from video and text is the possibility of performing video summarization with natural language~\cite{plummer2017enhancing}. Other works also tackle the problem of visual grounding of sentences: it can be applied to spatial grounding in images~\cite{johnson16densecap,plummer2015flickr30k,plummer2017phrase} or temporal grounding (\textit{i.e} temporal localization) in videos~\cite{bojanowski15weakly,hendricks17localizing}. Our method improves text-video embeddings and has potential to improve any method relying on such representations.

\subsection{Multi-stream video representation}

Combining different modalities is a straightforward way to improve video representations for many tasks. Most state-of-the-art video representations~\cite{feichtenhofer16convolutionaltwostream,feichtenhofer17spatiotemporalmultiplier,simonyan2014,miech17learnable,wang16temporal} separate videos into multiple stream of modalities. The appearance, which are features capturing visual cues, the motion, computed from optical flow estimation or dense trajectories~\cite{wang13action}, and the audio signal are the commonly used video modalities.
Investigating on which video descriptors to combine and how to efficiently fuse them
has been extensively studied.
Most prior works~\cite{carreira2017quovadis,feichtenhofer16convolutionaltwostream,girdhar17actionvlad,simonyan2014,varol17longterm} address the problem of appearance and motion fusion for video representation.
Other more recent works~\cite{arandjelovic17look,miech17learnable} explore appearance-audio two-stream architectures for video representation. This and other work has consistently demonstrated the benefits of combining different video modalities for tasks such as video classification and action recognition. Similar to previous work in video understanding, our model combines multiple modalities but can also handle missing modalities during training and testing.

\subsection{Learning with missing data}
Our work is also closely related to learning methods designed to handle missing data. Handling missing data in machine learning is far from being a solved problem, yet it is widespread in various fields. Data can be missing due to several reasons: it can be corrupted, it may have not been possible to record the data or, in some cases, the data may be intentionally missing (take an example of forms with answers to some fields being optional). Common practices in machine learning aim at imputing the missing values with a default value such as zero, the mean, the median or the most frequent value in the discrete case\footnote{\url{http://scikit-learn.org/stable/modules/generated/sklearn.preprocessing.Imputer.html}}. In the matrix completion theory, a low rank approximation of the matrix~\cite{jain2013low} can be performed to fill the missing values. In computer vision, one main application of learning with missing data is the inpainting task. Several approaches such as: Low rank matrix factorization~\cite{mairal10online}, Generative Adversarial Network~\cite{yeh17semantic} or more recently~\cite{ulyanov17deepimageprior} have successfully addressed the problem. The UberNet network~\cite{kokkinos17ubernet} is a universal multi-task model aiming at solving multiple problems such as: object detection, object segmentation or surface normal estimation. To do so, the model is trained on a mix of different annotated datasets, each one having its own task-oriented set of annotation. Their work is also related to ours as we also combine diverse types of datasets. However in our case, we have to address the problem of missing video modalities instead of missing task annotation.

Handling missing modalities can be seen as a specific case of learning from missing data. In image recognition the recent work~\cite{tran17missingmodalities} has tackled the task of learning with missing modalities to treat the problem of missing sensor information. In this work, we address the problem of missing video modalities. As explained above, videos can be divided into multiple relevant modalities such as appearance, audio and motion. Being able to train and infer models without all modalities makes it possible to mix different type of data such as illustrated in 
Figure~\ref{fig:teaser}.

%\section{Mixture-of-Embedding-Experts}
%\section{Modelling video and text with a mixture of embedding experts}
%%%%%%%%%%%%%%%%%%%%%%%%%%%%%%%%
%
\section{Mixture of embedding experts for video and text}
%
%%%%%%%%%%%%%%%%%%%%%%%%%%%%%%%%%%%%%
%In this section we describe our proposed mixture of embedding experts model (MEE) for heterogeneous data sources with incomplete input streams.
% during both training and inference.

%We introduce in this section our proposed Mixture-of-Embedding-Experts (MEE) model.
%Moreover, we will explain how our model can properly handle heterogeneous input sources with incomplete sets of data streams during both training and inference.

\begin{figure}[t]
  \begin{center}
     \includegraphics[width=\textwidth]{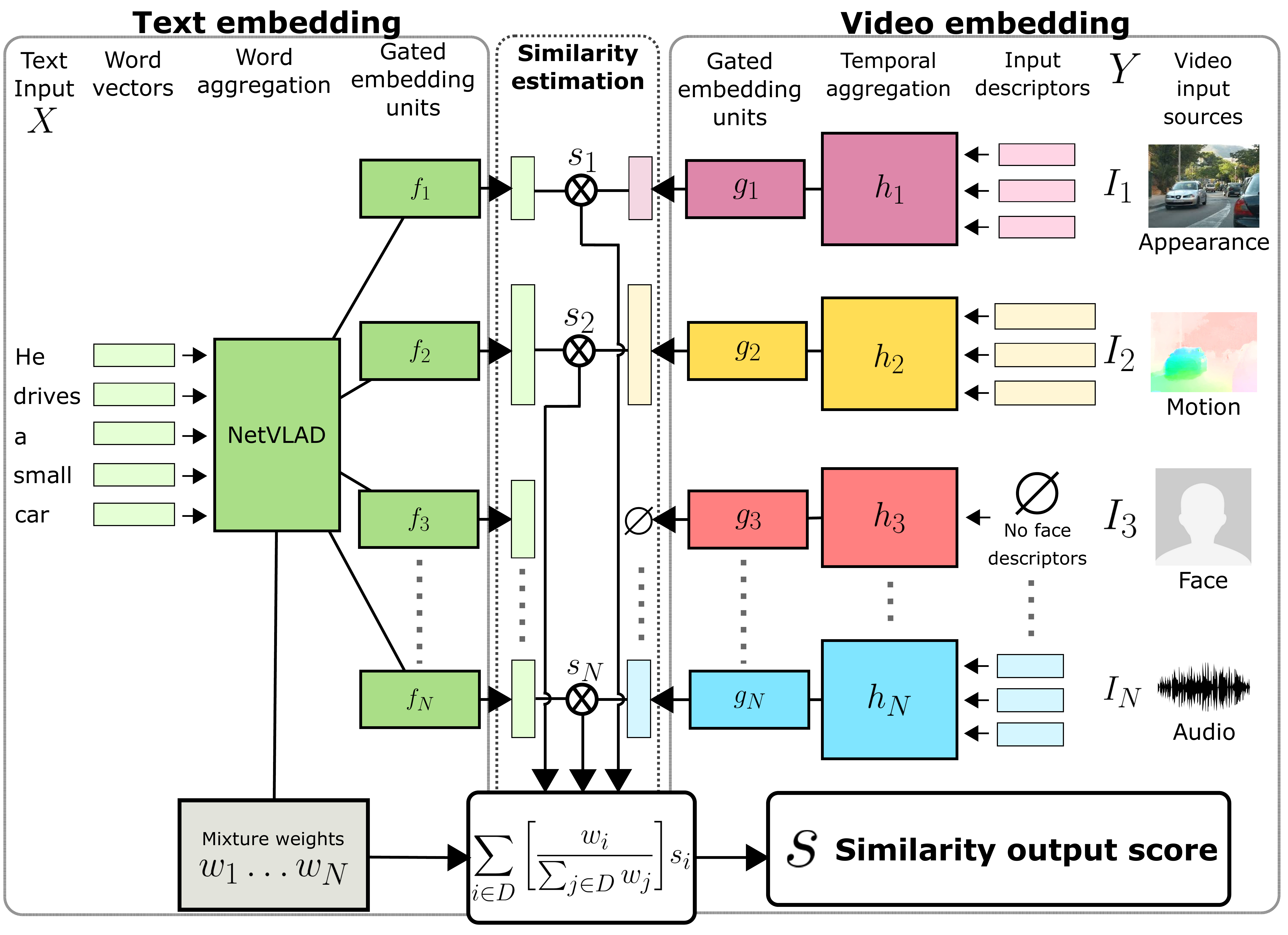}
\end{center}
\vspace{-0.5cm}
\caption{\small{\bf Mixture of embedding experts (MEE) model} that computes similarity score $s$ between input sentence $X$ and video $Y$ as a weighted combination of expert embeddings, one for each input descriptor type including appearance, motion, facial descriptors or audio. The appropriate weight of each expert is estimated from the input text. 
Our model can deal with missing video input such as face descriptors missing for videos without people depicted above.
}
\label{fig:separate_embd}
\end{figure}

In this section we introduce the proposed mixture of embedding experts (MEE) model and explain how this model handles heterogeneous input sources with incomplete sets of data streams during both training and inference.

\subsection{Model overview and notation}
Our goal is to learn a common embedding space for video and text.
More formally, if $X$ is a sentence and $Y$ a video, we would like to learn embedding functions $f$ and $g$ such that similarity $ s = \langle f(X), g(Y) \rangle$ is high if and only if $X$ and $Y$ are semantically similar. 
We assume that each input video is composed of $N$ different streams of descriptors, $\{I_{i}\}_{i \in 1\ldots N}$ that represent, for example, motion, appearance, audio, or facial appearance of people. Note that as we assume the videos come from diverse data sources a particular video may contain only a subset of these descriptor types. For example, some videos may not have audio, or will not have face descriptors when they don't depict people. As we will show later, the same model will be able to represent still images as (very) simple videos composed of a single frame without motion. To address the issue that not all videos will have all descriptors, we design a model inspired by the mixture of experts~\cite{jordan94mixture}, where we learn a {\em separate} ``expert" embedding model for each descriptor type. The expert embeddings are combined in an end-to-end trainable fashion using weights that depend on the input caption. As a result, the model can learn to increase the relative weight of motion descriptors for input captions concerning human actions, or increase the relative weight of face descriptors for input captions that require detailed face understanding.

The overview of the model is shown in Figure~\ref{fig:separate_embd}. 
Descriptors of each input stream $I_i$ are first aggregated over time using the temporal aggregation module $h_i$ and the resulting aggregated descriptor is embedded using a gated embedding module $g_i$ (see \ref{sec:gated-embd}).  Similarly, the individual word embeddings from the input caption are first aggregated using a text aggregation module into a single descriptor, which is then embedded using gated embedding modules $f_i$, one for each input source $i$. The resulting expert embeddings for each input source are then weighted using normalized weights $w_i(X)$ estimated by the weight estimation module from caption $X$ to obtain the final similarity score $s$. Details of the individual components are given next.

\subsection{Text representation}

The textual input is a sequence of word embeddings for each input sentence. These individual word embedding vectors are then aggregated into a single vector representing the entire sentence using a NetVLAD~\cite{arandjelovic16netvlad} aggregation module, denoted $h(X)$.  This is motivated by the recent results~\cite{miech17learnable} demonstrating superior performance of NetVLAD aggregation over other common aggregation architectures such as long short-term memory (LSTM)~\cite{hochreiter97lstm} or gated recurrent units (GRU)~\cite{cho11GRU}.

\subsection{Temporal aggregation module}

Similar to input text, each input stream $I_i$ of video descriptors is first aggregated into a single vector using temporal aggregation module $h_i$. For this, we use NetVLAD~\cite{arandjelovic16netvlad} or max pooling, depending on the input descriptors. Details are given in Section~\ref{sec:exp}.

\subsection{Gated embedding module} \label{sec:gated-embd}
The gated embedding module $Z=f(Z_0)$ takes a $d_1$-dimensional feature $Z_0$ as input and embeds (transforms) it into a new feature $Z$ in $d_2$-dimensional output space.
This is achieved using the following sequence of operations:

\begin{align}
 &Z_{1} = W_1Z_{0}+b_1, \label{eq:gem1}\\ 
 &Z_{2} = Z_{1} \circ \sigma(W_2Z_{1}+b_2), \label{eq:gem2} \\
 &Z = \frac{Z_{2}}{\|Z_{2} \|_{2}}, \label{eq:gem3}
\end{align}

where $W_1 \in \mathbb{R}^{d_2 \times d_1}, W_2 \in \mathbb{R}^{d_2 \times d_2}, b_1 \in \mathbb{R}^{d_2}, b_2 \in \mathbb{R}^{d_2} $ are  learnable parameters, $\sigma$ is an element-wise sigmoid activation and $\circ$ is the element-wise multiplication (Hadamard product). %Please note that equation~\eqref{eq:gem1}
Note that the first layer, given by~\eqref{eq:gem1}, describes a projection of the input feature $Z_{0}$ to the embedding space $Z_{1}$. The second layer, given by~\eqref{eq:gem2}, performs context gating~\cite{miech17learnable}, where individual dimensions of $Z_{1}$ are reweighted using learnt gating weights $ \sigma(W_2Z_{1}+b_2)$ with values between 0 and 1, where  $W_2$ and $b_2$ are learnt parameters. The motivation for such gating is two-fold: (i) we wish to introduce non-linear interactions among dimensions of $Z_{1}$ and (ii) we wish to recalibrate the strengths of different activations of $Z_{1}$ through a self-gating mechanism.
Finally, the last layer, given by~\eqref{eq:gem3}, performs L2 normalization to obtain the final output $Z$.

\subsection{Estimating text-video similarity with a mixture of embedding experts}
In this section we explain how to compute the final similarity score between the input text sentence $X$ and video $Y$. Recall, that each video is represented by several input streams $I_i$ of descriptors. % denoted $I_{1}, \dots, I_{N}$. 
Our proposed model learns separate (expert) embedding between the input text and each of the input video streams. These expert embeddings are then combined together to obtain the final similarity score. More formally, we first compute a similarity score $s_i$ between the input sentence $X$ and input video stream $I_i$
\begin{align}
 s_{i}(X,I_i) = \langle f_{i}(h(X)), g_{i}(h_i(I_{i})) \rangle,
\end{align}
where $f_{i}(h(X))$ is the text embedding composed of aggregation module $h()$ and gated embedding module $f_i()$;  $g_{i}(h_i(I_{i}))$ is the embedding of the input video stream $I_i$ composed of descriptor aggregation module $h_i$ and gated embedding module $g_i$; and $ \langle a, b \rangle$ denotes a scalar product.  Please note that we learn a separate text embedding $f_i$ for each input video stream $i$. In other words, we learn different embedding parameters to match the same input sentence $X$ to different video descriptors. For example, such embedding can learn to emphasize words related to facial expressions when computing similarity score between the input sentence and the input face descriptors, or to emphasize action words when computing the similarity between the input text and input motion descriptors.

\paragraph{Estimating the final similarity score with a mixture of experts.}

The goal is to combine the similarity scores $s_i(X,I_i)$ between the input sentence $X$ and different streams of input descriptors $I_i$ into the final similarity score. To achieve that we employ the mixture of experts approach~\cite{jordan94mixture}. 
%In detail, we consider scores    $s_i(X,I_i)$ as independent  experts for different descriptor types. 
In detail, the final similarity score $s(X,Y)$ between the input sentence $X$ and video $Y$ is computed as 
\begin{align}
\label{eq:moe}
 s(X,Y) = \sum_{i = 1}^{N} w_i(X)s_i(X,I_i),  \text{with } \ \ w_i(X) = \frac{e^{h(X)^{\top} a_{i}}}{\sum_{j=1}^{N} e^{h(X)^{\top} a_{j}}},
\end{align}
where $w_i(X)$ is the weight of similarity score $s_i$ predicted from the input sentence $X$, $h(X)$ is the aggregated sentence representation and $a_i$, $i=1\ldots N$ the learnt parameters. Please note again that the weights $w_i$ of experts $s_i$ are predicted from sentence $X$. In other words, the input sentence provides a prior on which of the embedding experts to put more  weight to compute the final global similarity score. The estimation of the weight of the different input streams can be seen as an attention mechanism that uses the input text sentence. 
%For example, if the caption talks about objects the weight can focus on appearance stream, whereas if the caption talks about facial expressions more emphasis can be placed on the facial appearance stream.  
For instance,  we may expect to have high weight on the motion stream for input captions such as: ``The man is practicing karate'', facial descriptors for captions such as ``Barack Obama is giving a talk'',  or on audio descriptors for input captions such as ``The woman is laughing out loud''.  

\paragraph{Single text-video embedding.}
Please note that equation \eqref{eq:moe} can be viewed as a single text-video embedding $s(X,Y) = \langle f(X), g(Y) \rangle$, where:\\ $f(X) = [w_1(X)f_{1}(h(X)), \dots, w_N(X)f_{N}(h(X))]$ is the vector concatenating individual text embedding vectors $f_{i}(h(X))$ weighted by estimated expert weights $w_i$, and  $g(Y) = [g_{1}(h_1(I_1)), \dots, g_{N}(h(I_{N}))]$ is the concatenation of the individual video embedding vectors $g_i(h_i(I_i))$. This is important for retrieval applications in large-scale datasets, where individual embedding vectors for text and video can be pre-computed offline and indexed for efficient search using techniques such as product quantization~\cite{jegou11product}.

\paragraph{Handling videos with incomplete input streams.} 
The formulation of the similarity score $s(X,Y)$ as a mixture of experts provides a proper way to handle situations where the input set of video streams is incomplete. For instance, when audio descriptors are missing for silent videos or when face descriptors are missing in shots without people.  In detail, in such situations we estimate the similarity score $s$ using the remaining available experts by renormalizing the remaining mixture weights to sum to one as  
\begin{align}
  s(X,Y) = \sum_{i \in D} \bigg[ \frac{w_i(X)}{\sum_{j \in D} w_j(X)}\bigg]s_i(X,I_i), 
\end{align}
 where $D \subset \{1\ldots N\}$ indexes the subset of available input streams $I_i$ for the particular input video $Y$. 
When training the model, the gradient thus only backpropagates to the available branches of both text and video.

\subsection{Bi-directional ranking loss}
To train the model, we use the bi-directional max-margin ranking loss~\cite{karpathy14deepfragment,wang2016learning,yu16videocaptioning,wang2018learning} as we would like to learn an embedding that works for both text-to-video and video-to-text retrieval tasks.
More formally, at training time, we sample a batch of sentence-video pairs $(X_{i},Y_{i})_{i \in [1,B]}$ where $B$ is the batch size. We wish to enforce that, for any given $i \in [1,B]$, the similarity score $s_{i,i} = s(X_{i}, Y_{i})$ between video $Y_i$ and its ground truth caption $X_i$ is greater than every possible pair of scores $s_{i,j}$ and $s_{j,i}$, where $j \neq i$ of non-matching videos and captions. This is implemented by using the following loss for each batch of $B$ sentence-video pairs $(X_{i},Y_{i})_{i \in [1,B]}$
\begin{align}
\label{loss}
l = \frac{1}{B}\sum_{i=1}^{B} \sum_{j \neq i} \Big[ \max(0, m + s_{i,j} - s_{i,i}) + \max(0, m + s_{j,i} - s_{i,i})\Big],
\end{align}
where $s_{i,j} = s(X_{i},Y_{j})$ is the similarity score of sentence $X_i$ and video $Y_j$, and
 $m$ is the margin. We set $m = 0.2$ in practice.

%%%%%%%%%%%%%%%%%%%%%%
\section{Experiments} \label{sec:exp}
%%%%%%%%%%%%%%%%%%%%%%

In this section, we report experiments with our mixture of embedding experts (MEE) model on different text-video retrieval tasks. We perform an ablation study to highlight the benefits of our approach and compare the proposed model with current state-of-the-art methods. % on the LSMDC retrieval tasks.

\subsection{Experimental setup}
In the following, we describe the used datasets and details of data pre-processing and training procedures. 
%We describe here the used datasets, the data pre-processing and training details.

\paragraph{Datasets.} We perform experiments on the following three datasets:\\
{\noindent {\bf{1 - MPII movie description/LSMDC dataset.}}}
We report results on the MPII movie description dataset~\cite{rohrbach15dataset}. This dataset contains 118,081 short video clips extracted from 202 movies. Each video has a caption, either extracted from the movie script or from transcribed audio description. The dataset is used in the Large Scale Movie Description Challenge (LSMDC). We report experiments on two LSMDC challenge tasks: movie retrieval and movie annotation. The first task evaluates text-to-video retrieval: given a sentence query, retrieve the corresponding video from 1,000 test videos. The performance is measured using recall@k (higher is better) for different values of k, or median rank (lower is better).   
The second, movie annotation task evaluates video-to-text retrieval: we are provided with 10,053 short clips, where each clip comes with five captions, with only one being correct. The goal is to find the correct one. The performance is measured using the accuracy. For both tasks we follow the same evaluation protocol as described on the LSMDC website\footnote{\url{https://sites.google.com/site/describingmovies/lsmdc-2017}}.

{\noindent \bf{2 - MSR-VTT dataset.}}
We also report several experiments on the MSR-VTT dataset~\cite{xu16msrvtt}. This dataset contains 10,000 unique Youtube video clips. Each of them is annotated with 20 different text captions, which results in a total of 200,000 unique video-caption pairs. Because we are only provided with URLs for each video, some of the video are, unfortunately, not available for download anymore. In total, we have successfully downloaded 7,656 videos (out of the original 10k videos). Similar to the LSMDC challenge and~\cite{rohrbach15dataset}, we evaluate on the MSR-VTT dataset the text-to-video retrieval task on randomly sampled 1,000 video-caption pairs from the test set. 

{\noindent \bf{3 - COCO 2014 Image-Caption dataset.}} We also report results on the text to still image retrieval task on the 2014 version of the COCO image-caption dataset~\cite{lin14coco}. Again, we emulate the LSMDC challenge and evaluate text-to-image retrieval on randomly sampled 1000 image-caption pairs from the COCO 2014 validation set.
%To evaluate our models on the task of text-to-image retrieval, in the same way as it is performed in the LSMDC challenge test split, we randomly sample 1000 image-caption pairs from the COCO 2014 validation set.
\begin{table}[t]
  \setlength{\tabcolsep}{3pt}
      \caption{\small Ablation study on the MPII movie dataset. %\textit{COCO} stands for COCO data augmentation, \textit{Face} for the face features embedding and \textit{Gated Embd} denotes the use of our gated embedding units from subsection~\ref{sec:gated-embd} instead of simple linear projections.
      R@k denotes recall@k (higher is better), MR denotes mean rank (lower is better). Multiple choice is measured in accuracy (higher is better).
      }
    \centering
    \scalebox{1.0}{
    \begin{tabular}{@{}lcccc|c@{}}
        \toprule
       Evaluation task & \multicolumn{4}{c|}{Text-to-Video retrieval} & Video-to-Text retrieval\\  
      \midrule
      Method                   & R@1 & R@5 & R@10 & MR & Multiple Choice\\
      \midrule
      MEE               & $\textbf{10.2}$ & $25.0$ & $33.1$ & $29$ & $74.0$ \\
      MEE + Face                & $9.3$& $25.1$ & $33.4$  & $ \textbf{27}$ & $\textbf{74.8}$ \\
      MEE + COCO                 & $9.8$& $\textbf{25.6}$ & $\textbf{34.7}$ & $\textbf{27}$ & $73.4$ \\
      MEE + COCO + Face                & $10.1$& $\textbf{25.6}$ & $34.6$  & $ \textbf{27}$ & $73.9$ \\
      \bottomrule
    \end{tabular}
    }
      \label{table:mpii-ablation}
\end{table}

%\subsection{Input processing}
\paragraph{Data pre-processing.}
For text pre-processing, we use the Google News\footnote{GoogleNews-vectors-negative300} trained word2vec word embeddings~\cite{mikolov13efficient}. For sentence representation, we use NetVLAD \cite{arandjelovic16netvlad} with 32 clusters. %and limit each sentence to 30 word embeddings.
%
%{\noindent \bf{Video.} } 
For videos, we extract frames at 25 frames per seconds and resize each frame to have a consistent height of 300 pixels. 
We consider up to four different descriptors representing the visual appearance, motion, audio and facial appearance.
We pre-extract the descriptors for each input video resulting in up to four input streams of descriptors. 
%Thus for each them, our model is fed with a temporal stream of descriptors.
The appearance features are extracted using the Imagenet pre-trained ResNet-152~\cite{he16resnet} CNN. We extract 2048-dimensional features from the last global average pooling layer. %The aggregation function of the appearance features is a simple max pooling. 
The motion features are computed using a Kinetics pre-trained I3D flow network~\cite{carreira2017quovadis}. We extract the 1024-dimensional features from the last global average pooling layer. 
%The aggregation function of the motion features is also a max pooling. 
The audio features are extracted using the audio CNN~\cite{hershey17cnn}. 
%To aggregate the audio features, we use a NetVLAD module with 16 clusters. 
Finally, for the face descriptors, we use the dlib framework\footnote{\url{http://dlib.net/}} to detect and align faces. Facial features are then computed on the aligned faces using the same framework, which implements a ResNet CNN trained for face recognition. For each detected face, we extract 128-dimensional representation. 
%To aggregate these features across the video, we use a max pooling.
We use max-pooling operation to aggregate appearance, motion and face descriptors over the entire video. To aggregate the audio features, we follow~\cite{miech17learnable} and use a NetVLAD module with 16 clusters. 

%\subsection{Training details}
\paragraph{Training details.}
Our work was implemented using the PyTorch\footnote{\url{http://pytorch.org/}} framework. We train our models using the ADAM optimizer~\cite{kingma15adam}. On the MPII dataset, we use a learning rate of $0.0001$ with a batch size of 512. On the MSR-VTT dataset, we use a learning rate of $0.0004$ with a batch size of 64. Each training is performed using a single GPU and takes only several minutes to finish. % is rather fast: \textit{several minutes are required at most}.

%\subsection{Ablation experiments}
\subsection{Benefits of learning from heterogeneous data}

The proposed embedding model is designed for learning from diverse and incomplete inputs. We demonstrate this ability on two examples. First, we show how a text-video embedding model can be learnt by augmenting captioned video data with captioned still images. For this we use the Microsoft COCO dataset~\cite{lin14coco} that contains captions provided by humans. Methods augmenting training data with still images from the COCO dataset are denoted (+COCO). Second, we show how our embedding model can incorporate an incomplete input stream of facial descriptors, where face descriptors are present in videos containing people but are absent in videos without people. Methods that incorporate face descriptors are denoted (+Face).

\paragraph{Ablation study on the MPII movie dataset.}
Table~\ref{table:mpii-ablation} shows a detailed ablation study on the LSMDC Text-to-Video and Video-to-Text retrieval tasks on the MPII movie dataset. 
Unfortunately here, we notice that incorporating heterogeneous data does not seem to significantly help the retrieval performances.

\begin{table}[t]
  \setlength{\tabcolsep}{3pt}
      \caption{\small The effect of augmenting the MPII movie caption dataset with captioned still images from the MS COCO dataset. R@k denotes recall@k (higher is better), MR denotes Median Rank (lower is better) and MC denotes Multiple Choice (higher is better).}
    \centering
    \scalebox{0.9}{
    \begin{tabular}{@{}l|cccc|ccccc@{}}
      \hline
       Evaluation set & \multicolumn{4}{c|}{COCO images} & \multicolumn{5}{c}{MPII videos}\\    
      \toprule
      Model                  & R@1 & R@5 & R@10 & MR & R@1 & R@5 & R@10 & MR & MC\\
      \midrule
      MEE + Face             & $10.4$ & $29.0$ & $42.6$ & $15$ & $9.3$& $25.1$ & $33.4$  & $ \textbf{27}$ & $\textbf{74.8}$\\
      MEE + Face + COCO          & $\textbf{31.4}$& $\textbf{64.5}$ & $ \textbf{79.3}$ & $\textbf{3}$ & $\textbf{10.1}$& $\textbf{25.6}$ & $\textbf{34.6}$  & $ \textbf{27}$ & $ 73.9$\\
      \bottomrule
    \end{tabular}
    }
      \label{table:coco-comparison}
\end{table}

\begin{table}[t]
  \setlength{\tabcolsep}{3pt}
      \caption{\small The effect of augmenting the MSR-VTT video caption dataset with captioned still images from the MS COCO dataset when relative image to video sampling rate $\alpha = 0.5$. R@k stands for recall@k, MR stands for Median Rank.}
    \centering
    \scalebox{0.95}{
    \begin{tabular}{@{}l|cccc|cccc@{}}
      \hline
       Evaluation set & \multicolumn{4}{c|}{COCO images} & \multicolumn{4}{c}{MSR-VTT videos}\\    
      \toprule
      Model                  & R@1 & R@5 & R@10 & MR & R@1 & R@5 & R@10 & MR\\
      \midrule
      MEE + Face             & $8.4$ & $24.9$ & $38.9$ & $18$ & $13.6$& $37.9$ & $51.0$ & $10$\\
      MEE + Face + COCO          & $\textbf{20.7}$& $\textbf{54.5}$ & $ \textbf{72.0}$ & $\textbf{5}$ & $\textbf{14.2} $& $ \textbf{39.2}$ & $ \textbf{53.8}$ & $\textbf{9}$ \\
      \bottomrule
    \end{tabular}
    }
      \label{table:msrvtt-comparison}
\end{table}

\paragraph{Augmenting videos with images.}

Next, we evaluate in detail the benefits of augmenting captioned video datasets (MSR-VTT and MPII movie) with captioned still images from the Microsoft COCO dataset. 
%First, we evaluate the effect of augmenting the MPII movie description. 
%Because of the scale of the COCO training set is roughly equivalent to the size of the MPII movie description dataset, we did not use any sampling strategy when mixing both datasets.
 Table~\ref{table:coco-comparison} shows the effect of adding the still image data during training. For all models, we report results on both the COCO image dataset and the MPII videos. Adding COCO images to the video training set improves performance on both COCO images but also MPII videos, showing that a single model trained from the two different data sources can improve performance on both datasets.
This is an interesting result as the two datasets are quite different in terms of depicted scenes and textual captions.
MS COCO dataset contains mostly Internet images of scenes containing multiple objects. MPII dataset contains video clips from movies often depicting people interacting with each other or objects.

\begin{figure}[t]
  \begin{center}
     \includegraphics[width=1.05\textwidth]{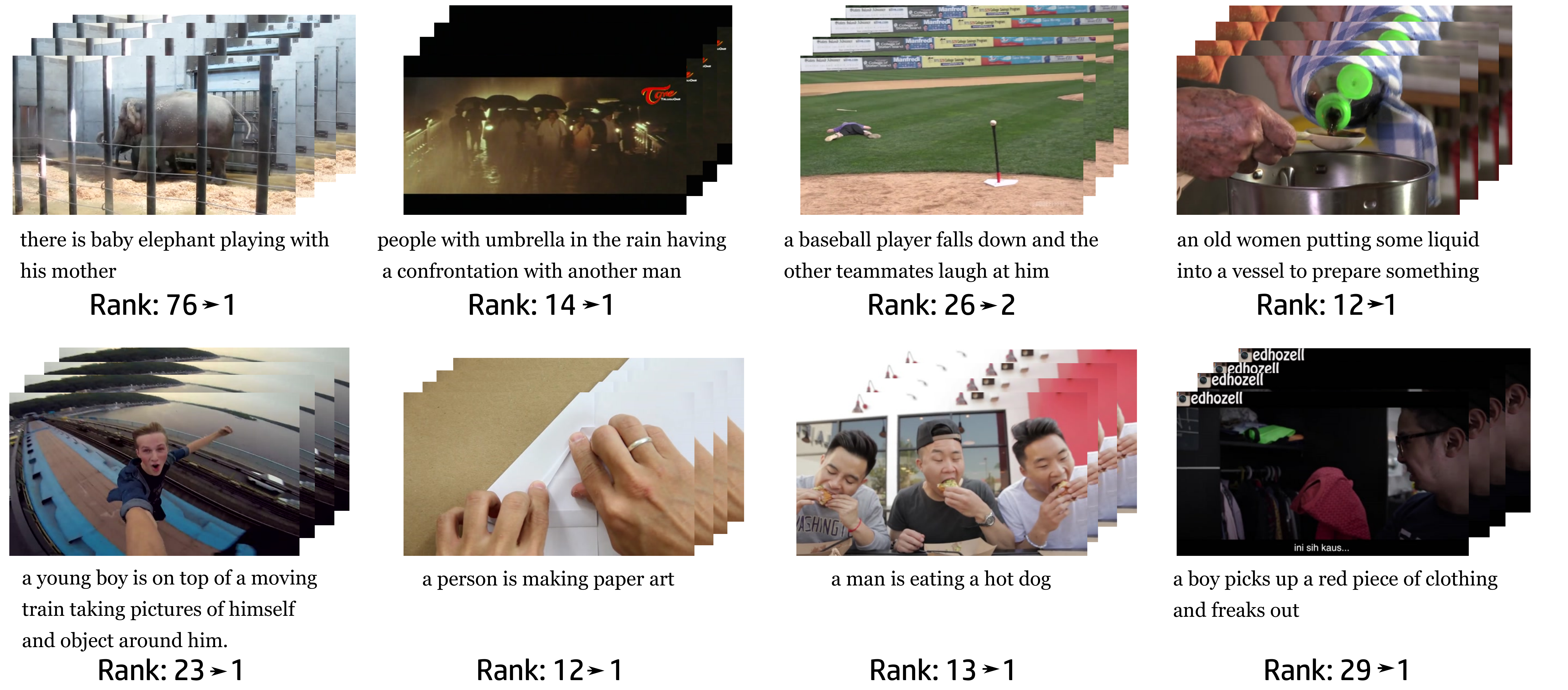}
\end{center}
\vspace{-0.6cm}
\caption{\small Example videos with large relative improvement in text-to-video retrieval ranking (out of 1000 test videos) on the MSR-VTT dataset when incorporating still images from the COCO dataset at training using our proposed MEE model. Notice that the improved videos involve querying objects frequently present in the COCO dataset including: \textit{elephant}, \textit{umbrella}, \textit{baseball} or  \textit{train}.}
\label{fig:msr-viz}
\end{figure}

We also evaluate the impact of augmenting MSR-VTT video caption dataset with the captioned still images from the MS COCO dataset. As the MSR-VTT is much smaller than the COCO dataset, it becomes crucial to carefully sample COCO image-caption samples when augmenting MSR-VTT during training. In detail, for each epoch,  we randomly inject image-caption samples such that the ratio of image-caption samples to video-caption samples is set to a fixed sampling rate: $\alpha \in \mathbb{R}_{\ge 0}$. Note that $\alpha = 0$ means that no data augmentation is performed and $\alpha = 1.0$ means that exactly the same amount of COCO image-caption and MSR-VTT video-caption samples are used at each training epoch. 
Table~\ref{table:msrvtt-comparison} shows the effect of still image augmentation on the MSR-VTT dataset for the text-to-video retrieval task when the proportion of image-caption samples is half of the MSR-VTT video caption samples, i.e. $\alpha=0.5$. 
Our proposed MEE model fully leverages the additional still images. Indeed, we observe gains in video retrieval performances for all metrics. Figure~\ref{fig:msr-viz} shows qualitative results of our model highlighting some of the best relative improvement in retrieval ranking using the still image data augmentation. Note that many of the improved queries involve objects  frequently appearing in the COCO dataset including \textit{elephant}, \textit{umbrella}, \textit{baseball} or  \textit{train}.

% *** perhaps say some conclusion about choosing \alpha

\begin{table}[t]
  \setlength{\tabcolsep}{3pt}
    \centering
        \caption{\small Text-to-video and Video-to-Text retrieval results from the LSMDC test sets. MR stands for Median Rank, MC for Multiple Choice.}
    \scalebox{1.0}{
    \begin{tabular}{@{}lcccc|c@{}}
        \toprule
       Evaluation task & \multicolumn{4}{c|}{Text-to-Video retrieval} & Video-to-Text retrieval\\  
      \midrule
      Method                   & R@1 & R@5 & R@10 & MR & MC Accuracy\\
      \midrule
      Random baseline   & $0.1$ & $0.5 $ & $1.0 $ & $500$ & $20.0$ \\
      C+LSTM+SA+FC7~\cite{torabi16learning}      & $4.2 $ & $13.0 $ & $19.5 $ & $90$ & $58.1 $\\
      SNUVL~\cite{yu16videocaptioning} (LSMDC16 Winner)   & $3.6 $ & $14.7 $ & $23.9 $ & $50$ & $65.7 $ \\
      CT-SAN~\cite{yu17endtoend}    & $5.1 $ & $16.3 $ & $25.2 $ & $46$ & $67.0 $\\
      Miech \textit{et al.}~\cite{miech17learning}    & $7.3 $ & $19.2 $ & $27.1 $ & $52$ & $69.7 $\\
      CCA (FV HGLMM)~\cite{klein15associating} (\textit{same features})     & $7.5 $ & $ 21.7$ & $ 31.0$ & $33$ & $ 72.8$\\
      JSFusion~\cite{yu17jsf} (LSMDC17 Winner)   & $9.1$ & $21.2$ & $34.1$ & $36$ & $73.5$ \\
      \midrule
      MEE + COCO + Face (Ours)             & $\textbf{10.1} $& $ \textbf{25.6}$ & $\textbf{34.6}$ & $\textbf{27}$ & $\textbf{73.9}$ \\
      \bottomrule
    \end{tabular}
    }
      \label{table:mpii-comparison}
\end{table}

\subsection{Comparison with state-of-the-art}
Table~\ref{table:mpii-comparison} compares our best approach to the state-of-the-art results on the LSMDC challenge test sets. Note that our approach significantly outperforms all other available results including JSFusion~\cite{yu17jsf}, which is the winning method of the LSMDC 2017 Text-to-Video and Video-to-Text retrieval challenge. We also reimplemented the normalized CCA approach from Klein \textit{et al.}~\cite{klein15associating}. To make the comparison fair, we used our video features and word embeddings. %We also report the zero-padding baseline. 
Finally, we also significantly outperform the C+LSTM+SA+FC7~\cite{torabi16learning} baseline that augments the MPII movie dataset with COCO image caption data.

\section{Conclusions}
We have described a new model, called mixture of embedding experts (MEE), that learns text-video embeddings from heterogeneous data sources and is able to deal with missing video input modalities during training. We have shown that our model can be trained from image-caption and video-caption datasets treating images as a special case of videos without motion and sound. In addition, we have demonstrated that our model can optionally incorporate at training, input stream of facial descriptors, where faces are present in videos containing people but missing in videos without people. We have evaluated our model on the task of video retrieval. Our approach outperforms all reported results on the MPII Movie Description. Our work opens-up the possibility of learning text-video embedding models from large-scale weakly-supervised image and video datasets such as the Flickr 100M~\cite{thomee2016yfcc100m}.  % Cite: http://yfcc100m.appspot.com/

%\clearpage

\paragraph{Acknowledgments.}
The authors would like to thank Valentin Gabeur for spotting a bug in our codebase that affected multiple results from the initial paper.
The bug was fixed in the following commit: \url{https://tinyurl.com/s6hvn9s}.
This work has been partly supported by ERC grants ACTIVIA (no.\
307574) and LEAP (no.\ 336845), CIFAR Learning in Machines $\&$ Brains
program, European Regional Development Fund under the project IMPACT \\(reg. no.
CZ.02.1.01/0.0/0.0/15 003/0000468) and a Google Research Award.

\bibliographystyle{splncs}
\bibliography{master-biblio}
\end{document}